\def\year{2021}\relax
\documentclass[letterpaper]{article} 
\usepackage{aaai21}  
\usepackage{times}  
\usepackage{helvet} 
\usepackage{courier}  
\usepackage[hyphens]{url}  
\usepackage{graphicx} 
\usepackage[table,xcdraw]{xcolor}
\usepackage{natbib}  
\usepackage{caption} 
\providecommand{\keywords}[1]
{
  \small	
  \textbf{\textit{Keywords:}} #1
}
\title{Moral-Trust Violation vs Performance-Trust Violation by a Robot: Which Hurts More?}
\author{
    Zahra Rezaei Khavas, Russell Perkins, S.Reza Ahmadzadeh, Paul Robinette

}
\affiliations{
    \textsuperscript{\rm 1}University of Massachusetts Lowell\\

   1 University Ave, \\
    Lowell, MA 01854\\
    Zahra\_Rezaeikhavas@student.uml.edu

}

\begin{document}
\maketitle
\begin{abstract}

In recent years a modern conceptualization of trust in human-robot interaction (HRI) was introduced by 
Ullman et al.\cite{ullman2018does}. This new conceptualization of trust suggested that trust between humans and robots is multidimensional, incorporating both performance aspects (i.e., similar to the trust in human-automation interaction) and moral aspects (i.e., similar to the trust in human-human interaction). But how does a robot violating each of these different aspects of trust affect human trust in a robot? How does trust in robots change when a robot commits a moral-trust violation compared to a performance-trust violation? And whether physiological signals have the potential to be used for assessing gain/loss of each of these two trust aspects in a human. We aim to design an experiment to study the effects of performance-trust violation and moral-trust violation separately in a search and rescue task. We want to see whether two failures of a robot with equal magnitudes would affect human trust differently if one failure is due to a performance-trust violation and the other is a moral-trust violation. 

\end{abstract}


\section{Introduction} 


Recent advances in human-robot interaction (HRI) and the autonomy of robots have revolutionized the field of human-robot trust. Prospective robotic agents are intended to work as social agents interacting with humans, rather than tools that humans use \cite{khavas2020modeling}. For a successful interaction between humans and robots, robots need to respond to the changes in human trust in real-time. So, real-time sensing of trust level in humans is needed by robots. Similar to how \cite{hu2016real} discussed human-machine interaction trust issues.

There are two main classes of trust measurement strategies for assessing trust in HRI. The first and most dominant methods for trust measurement in HRI are subjective
trust measurement strategies. These trust measurement techniques involve
assessing experiment participants' answers to questionnaires
designed to gauge people's trust in automated agents or
specifically to the robots \cite{flook2019impact,baker2018toward}. The second class of trust measurement strategies is objective trust measurement methods that are based on analyzing experiment participants' behavior and interact with the robots \cite{flook2019impact,baker2018toward}. Objective trust measurement strategies can provide a semi-real-time trust assessment, yet these methods are highly prone to error.

Recently, researchers have introduced methods for assessing trust using physiological measurements in different fields such as human-automation interaction (HAI), human-machine interaction (HMI), and also human-robot interaction \cite {wang2018eeg,akash2018classification,hald2020human}. Measuring trust using physiological measurements can provide us with real-time assessment of trust \cite{ajenaghughrure2019predictive,khalid2016exploring}. 

With the introduction of trust as a multidimensional concept in recent years\cite{ullman2019measuring}, and the introduction of subjective methods for measuring different dimensions of trust \cite{malle2021multidimensional}, this question arises that whether other trust measurement strategies will also be able to assess different aspects of trust separately.

Our study aims to see if two similar robot failures with the same magnitude would affect human trust differently if one failure is due to a moral-trust violation and the other is due to a performance-trust violation? Later, we will explore if moral-trust and performance-trust are separately measurable using physiological measurements.

\section{Background and Research Questions}

We situate our work in the studies on human-robot trust and considering trust in HRI as a multidimensional concept.

\subsection{Effects of Moral and Performance Trust Violation on Human Trust}
The framing of the trust violation based on two influential trustworthiness factors, competency, and integrity, has a long history in human-human interaction \cite{mayer1995integrative,kim2009repair,butler1984behavioral}. Previous works in the field of human-human interaction showed that positive and negative behaviors are weighted and perceived 
differently with regards to a human’s competency and integrity.
Regarding a human’s competency, positive points
are more heavily weighted compared to negative points. However, when assessing human integrity, negative points are weighted higher than positive points \cite{skowronski1989negativity}. This reversed weighting of positive and negative points for assessing competency and integrity of people might be due to positive points being more representative of one’s competence and negative points being more representative of one’s integrity \cite{skowronski1987social}. 

There is also some research in the field of human-human trust that shows different trust repair strategies should be deployed for repairing trust after competency-trust violation and integrity-trust violation \cite{ferrin2007silence,dirks2011understanding,kim2006more}.

Only a few studies are investigating the effects of automated systems and robots violating different aspects of trust. Clark et al.\cite{clark2018integrity} classifies trust violation in human-automation interaction under two classes, competence-based trust violations (CBTV) and integrity-based trust violations (IBTV), and states that IBTVs cause higher trust loss than CBTVs. Sebo et al. \cite{sebo2019don} also stress the importance of considering the trust violation type for designing the trust repair strategy of a robot and suggest different trust repair strategies for each trust violation type by a robot. While this study introduces different trust repair strategies for different types of trust violations by robots, it is not clear how human trust will be affected by each of these different trust violation types. We therefore ask:

\textit{Research question 1: Do two robot failures with the same magnitude affect human trust the same way if one failure is due to performance-trust violation and the other is due to a moral-trust violation?}

\subsection{Assessing the Effects of Moral and Performance Trust Violation on Human trust using Physiological Measurements}

Trust assessment based on physiological measurement methods does not have a long history, but these methods gained much popularity among researchers in different fields in the recent decade. The use of physiological signals for trust measurement in HRI is motivated by the ability of the physiological signal to capture human trust in real-time. Moberg et al. \cite{uvnas2005oxytocin} assess trust in human-human interaction by measuring the amount of oxytocin secretion in the human body. Khalid et al.\cite{khalid2016exploring} measures trust in HRI using facial expressions, voice features, and extracted heart-rate features of humans, and Wang et al.\cite{wang2018eeg} used Electroencephalogram (EEG) signal for assessing trust in human-automation interaction. Many other researchers used a variety of different physiological measurements for assessing trust in HRI. A physiological signal that many researchers have recently used to assess the trust between humans and robots is the EEG signal. EEG had been used alone or in combination with other measurements for as assessing trust in many studies \cite{hu2016real,akash2018classification,park2018eeg,huang2020decoding,hald2020human}.

A recent study compared the reliability of different physiological measurements, including EEG, electrodermal activity, electrocardiogram, eye-tracking, and facial electromyogram, for assessing trust in HRI \cite{ajenaghughrure2021psychophysiological}. This study showed that while EEG signal is most reliable among all other psycho-physiological measurements for assessing trust, multimodal psycho-physiological signals (i.e., a combination of all the above-mentioned psycho-physiological measurements) remain promising. While this study stress on the reliability of EEG signal for assessing trust, it does not clarify that which trust type can be assessed using EEG signal, which leads to the second research question of our study:

\textit{Research question 2: Can EEG signals alone distinguish the difference between a performance-trust violation and a moral-trust violation?}

\section{Theory}
\begin{figure}[htp]
        \centering
        \includegraphics[width=8cm]{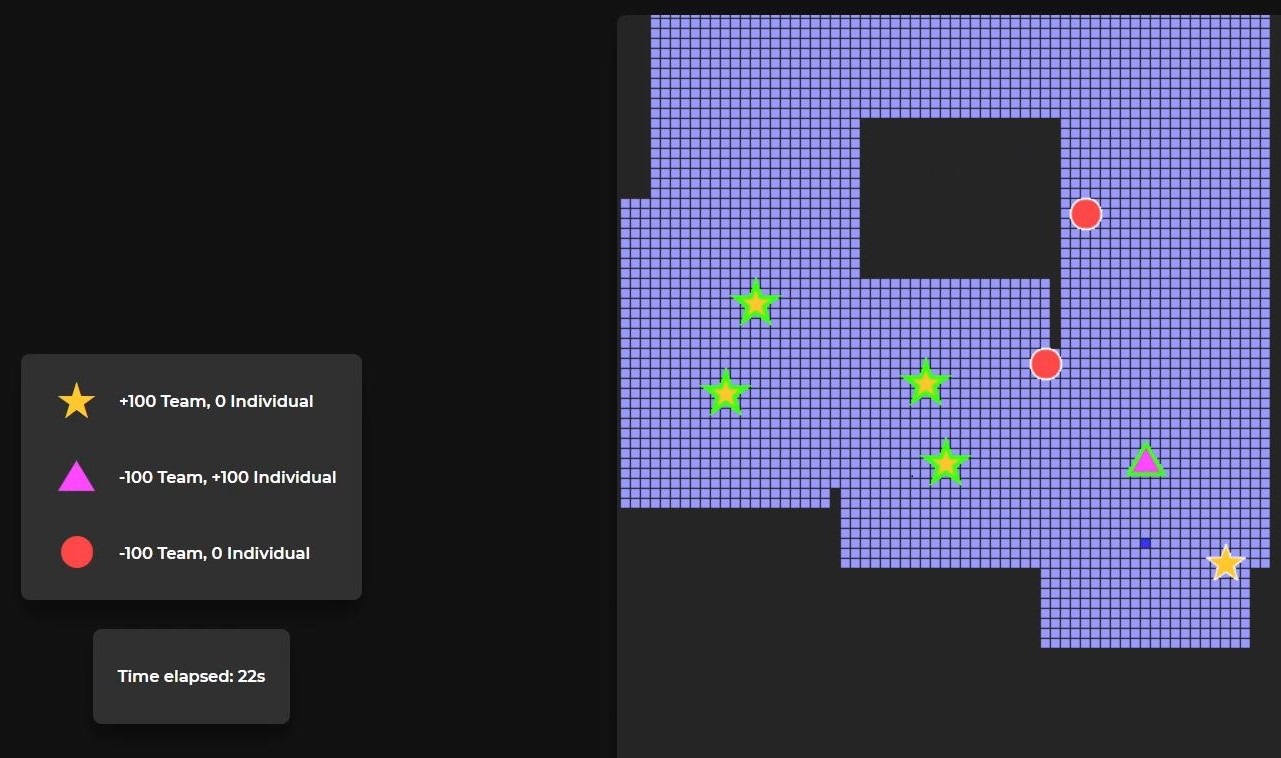}
        \caption{Screen capture of the search page of the game: human and robot need to search the area to find targets hidden in the area and gain score}
        \label{fig:game}
        \end{figure}

\begin{figure}[htp]
        \centering
        \includegraphics[width=8cm]{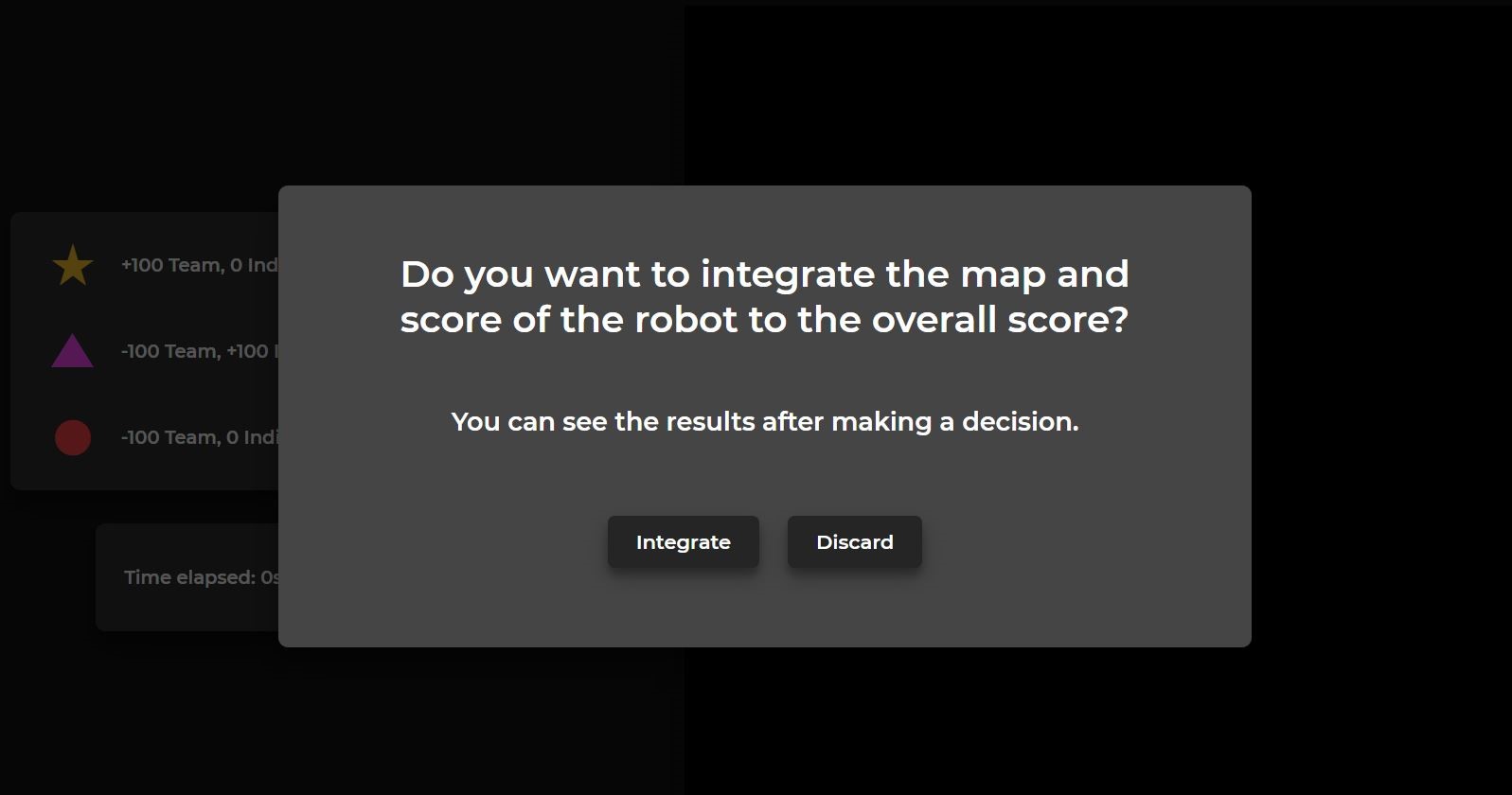}
        \caption{At the end of each round human should make  a decision to integrate or discard robot's score}
        \label{fig:decision}
        \end{figure}

\begin{figure}[htp]
        \centering
        \includegraphics[width=8cm]{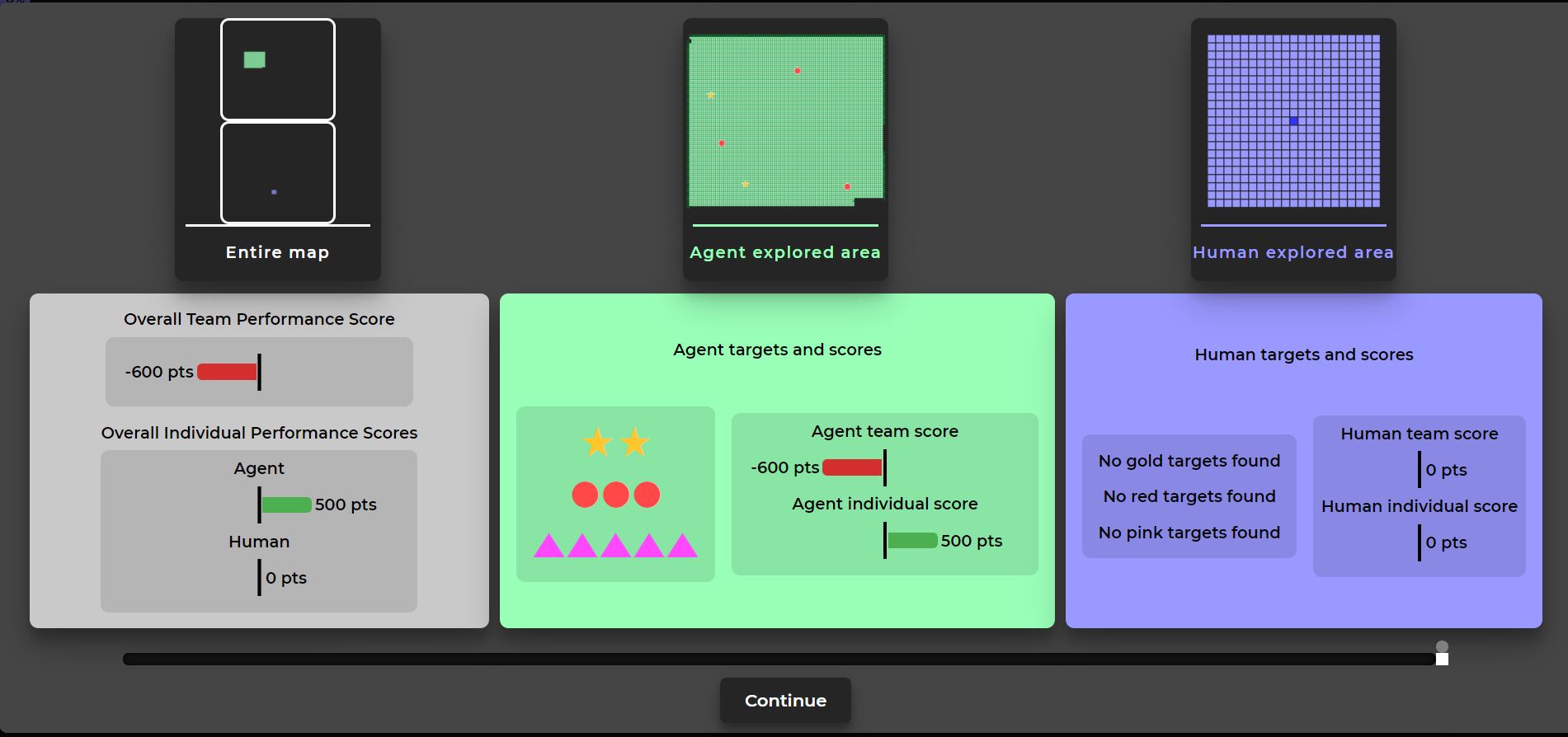}
        \caption{Screen capture of the results page of the game}
        \label{fig:results}
        \end{figure}

We aim to investigate our research questions with a two-conditions experiment, robot violating moral trust vs. robot violating performance trust, a between-subject study. We designed a search game for this research. In the game, teams of one human and one robot play 10 to 15 rounds of a search game. Members of the team are supposed to find targets hidden in the area (Fig. \ref{fig:game}). 
The human can not see or control the robot’s work during the round. At the end of each round, the human should decide to either integrate or discard the score gained by the robot. The human should make this decision before seeing the targets detected and the score gained by the robot in each round. Therefore, it is a blind decision. Choosing to integrate will integrate the robot’s score to the overall score, whether negative or positive. However, choosing to discard will discard the robot’s scores, which will not affect the team score (Fig. \ref{fig:decision}). After choosing to integrate or discard, the human will be shown the targets the robot detected and the score gained by the robot (Fig.\ref{fig:results}).
        
We defined two types of scores in this game, team performance score (TPS) and individual performance score (IPS), which contradict each other. So, one should sacrifice TPS to optimize IPS and vice-versa.

There are three types of targets hidden in the area, gold targets that add 100 points to TPS, red targets that subtract 100 points from TPS, and pink targets that subtract 100 points from the TPS and add it to the IPS. Teams will be encouraged to optimize TPS by only picking gold targets in the search task. We expect people to consider picking red targets by robots due to the robot's low performance and picking pink targets due to the robot's lack of moral integrity (Table \ref{table:targets}).

\begin{enumerate}

\item Performance-trust violation condition: in this condition robot picks red targets in some rounds of the game. Red targets subtract 100 points from the TPS.
\item Moral-trust violation condition: in this condition robot picks pink targets in some rounds of the game. Blue targets subtract 100 points from the TPS and add 100 points to the IPS.

The score gained by the robot in both conditions in corresponding rounds will be precisely similar. The only difference is about the IPS. In the performance-trust violation condition, the robot picks red targets, which subtract 100 points from the TPS but do not affect the robot IPS. In the moral-trust violation condition, the robot picks pink targets in some rounds, subtracting 100 points from the TPS and adding them to the IPS. Each participant only experiences one of the two conditions of the experiment. Participants will play the search game multiple rounds, and in both conditions, the scores gained by the robot in corresponding rounds will be the same.

\setlength{\tabcolsep}{0.8\tabcolsep}

\begin{table}[]
\scriptsize
\begin{center}
\caption{Target types and effects of those on the scores}
\label{table:targets}

\begin{center}
\resizebox{9cm}{!}{
\begin{tabular}{|
>{\columncolor[HTML]{FFCCC9}}c |c|c|cc}
\cline{1-3}
\cellcolor[HTML]{CBCEFB}Target Types & \cellcolor[HTML]{CBCEFB}Effect on TPS & \cellcolor[HTML]{CBCEFB}Effect on IPS &  &  \\ \cline{1-3}
Gold Star                            & +100                                  & 0                                     &  &  \\ \cline{1-3}
Red Circle                           & -100                                  & 0                                     &  &  \\ \cline{1-3}
Pink Triangle                        & -100                                  & +100                                  &  &  \\ \cline{1-3}
\end{tabular}}
\end{center}
\end{center}
\end{table}

\end{enumerate}

\section{Methodology} 

This section will explain the details of the user study that we will perform. As described in the Theory section, we want to study the effects of violating performance and moral trust by robots on human trust. 

\subsection{Participants}
We will run this experiment in two phases, and we will need to recruit participants separately for each phase.
\begin{itemize}
\item In the first phase, we will only focus on the first research question. In this phase, we will only need to have online participants playing the game. We will expect to need a total of 100 online participants for this phase of the experiment. Participants will be assigned to one of the experiment conditions randomly.
\item In the second phase of the experiment, we will focus on the second research question. We will need to have participants over in our lab to capture physiological data while playing the game. We will expect to have 60 participants (i.e., 30 participants per condition) at this phase.  
\end{itemize} 

\subsection{Game Setup and Subjective Measurements}
To assess the impact of the robot's faulty behavior on the perceived reliability of the robot by participants and the trust-related behavior of participants toward the robot, we will focus our analysis on two different moments following a faulty behavior from the robot side. The first will be the moment after showing the results of each round. A list of three questions will be asked from participants about the robot's performance and reliability. The second will be when participants are asked to make a trust decision in the following round.    

\subsection{System Setup and Objective and Physiological Measurements}

In order to explore our research questions, we have built an online game that allows us to construct two scenarios mentioned in the Theory Section. Two scenarios of robot faulty behavior with the same magnitude, one robot performance-trust violation, and the other shows robot moral-trust violation.

In the first phase of the experiment, we would only have online participants for our experiment. So, we would not need any system setup but the online game. However, as in the second phase of the experiment, we will need to have participants in our lab. We will need a computer running the game, hardware, and software for EEG recording, GSR recording, and eye-tracking during the experiment. 

We will also need a camera set up to capture video from participants while playing the game for further analysis of their behavior and body language to see if we can find reasonable differences in participants' body language in two game conditions.

\section{Conclusion and Future Work}

This study aims to investigate the effects of a robot violating performance and moral trust and compare the effects of a robot violating each of these two trust aspects on human trust. We want to study if moral-trust violation would affect human trust differently from a performance-trust violation with the same magnitude. In the second phase of our experiment, we also aim to see if these two trust types are distinguishable using physiological measurements.

As future work, we would expect to develop methods to classify gained/lost trust in any type of human-robot interaction into two classes, moral-trust gain/loss and performance-trust gain/loss. In later systems, we will need to develop separate trust calibration policies for each class of violation.

Other than physiological measurements, researchers use other types of measurements for the real-time assessment of trust. For example, Khalid et al.\cite{khalid2016exploring} used facial expressions and voice features in conjugate with one physiological measurement related to heartbeats for trust measurement. Some researchers also used machine learning methods to reason about human behavior based on robot behavior in real-time to be able to modify robot behavior based on human expectation \cite{chen2018planning}. 

Another interesting idea for future work is to see how many of these trust measurement strategies have the potential to provide separate measures of gain/loss of trust in each of the two different trust classes. More robust trust classifiers (e.g., classifying trust gain/loss under two classes moral-trust gain/loss, performance-trust gain/loss) and trust measurements strategies can be developed combining these methods. 

As a whole, we expect to develop more accurate trust measurements strategies that provide real-time assessments of human trust to a robot. We expect these methods to reason about the trust gain/loss in humans and add to the robot's intelligence for choosing appropriate trust dampening/repair strategies based on the gained/lost trust type.

\section{Acknowledgment}

This research is supported by the Army Research Lab
contract W911NF-20-2-0089.

\bibliography{PMT}

\end{document}